\documentclass[conference]{IEEEtran}
\IEEEoverridecommandlockouts
\usepackage{cite}
\usepackage{amsmath,amssymb,amsfonts}
\usepackage{algorithmic}
\usepackage{graphicx}
\usepackage{textcomp}
\usepackage{xcolor}
\definecolor{mycite}{HTML}{CC0099} 
\usepackage[
    colorlinks=true,
    linkcolor=blue,
    citecolor=mycite,
    urlcolor=black
]{hyperref}
\usepackage{multirow}
\usepackage{booktabs}
\usepackage{colortbl}
\usepackage{tabularx}
\usepackage[ruled,vlined]{algorithm2e}
\usepackage[table]{xcolor}
\usepackage[a4paper, total={184mm,239mm}]{geometry}
\def\BibTeX{{\rm B\kern-.05em{\sc i\kern-.025em b}\kern-.08em
    T\kern-.1667em\lower.7ex\hbox{E}\kern-.125emX}}

\begin{document}

\title{
\huge When Forgetting Builds Reliability: LLM Unlearning for Reliable Hardware Code Generation
}

\author
{
\IEEEauthorblockN{Yiwen Liang$^{\dagger}$,
Qiufeng Li$^{\dagger}$,
Shikai Wang,
Weidong Cao$^{*}$
}
\textit{Department of ECE, The George Washington University}, Washington, D.C., USA\\
\{yiwen.liang, qiufeng.li, shikai.wang,
weidong.cao\}@gwu.edu \\
$^{\dagger}$equal contribution, $^{*}$corresponding author
}
\IEEEaftertitletext{\vspace{-0.41cm}}
\maketitle

\begin{abstract}
Large Language Models (LLMs) have shown strong potential in accelerating digital hardware design through automated code generation. 
Yet, ensuring their reliability remains a critical challenge, as existing LLMs trained on massive heterogeneous datasets often exhibit problematic memorization of proprietary intellectual property (IP), contaminated benchmarks, and unsafe coding patterns. 
To mitigate these risks, we propose a novel unlearning framework tailored for LLM-based hardware code generation. 
Our method combines (i) a syntax-preserving unlearning strategy that safeguards the structural integrity of hardware code during forgetting, and (ii) a fine-grained floor-aware selective loss that enables precise and efficient removal of problematic knowledge.
This integration achieves effective unlearning without degrading LLM code generation capabilities. Extensive experiments show that our framework supports forget sets up to 3$\times$ larger, typically requiring only a single training epoch, while preserving both syntactic correctness and functional integrity of register-transfer level (RTL) codes.
Our work paves an avenue towards reliable LLM-assisted hardware design.

\end{abstract}

% \begin{IEEEkeywords}
% component, formatting, style, styling, insert
% \end{IEEEkeywords}

\section{Introduction}
\label{sec:intro}

% \textcolor{red}{To Yiwen: combine current paragraphs 1 and 2. Key points to cover: (a) Code generation capabilities of LLMs have been increasingly extended to hardware code generation; show some examples; (b) Yet, hardware code generation with LLMs has some fundamental challenges, e.g., security and trustworthy issues. (c) Show evidence of prior work to support your points and raise the key research questions. Overall, make this paragraph more concise.}

Recent advances in Large Language Models (LLMs) have ushered in a new paradigm for hardware development.
General-purpose models such as GPT, and domain-specific models like RTLCoder~\cite{liu2024rtlcoder} and VeriGen~\cite{thakur2024verigen}, have demonstrated remarkable capabilities in generating hardware codes, thereby substantially accelerating digital design workflows, shortening development cycles, and reducing manual engineering efforts.
\textbf{Despite this progress, fundamental challenges remain to ensure the reliability of LLM-based code generation. }
Trained on massive and heterogeneous datasets, current LLMs often memorize undesirable content, including proprietary intellectual property (IP), contaminated benchmarks, and unsafe coding patterns, which compromise evaluation fairness, risk IP leakage, and threaten the reliability of downstream hardware systems~\cite{dong2024generalization}. 
For example, benchmark frameworks such as VerilogEval~\cite{liu2023verilogeval} and Rtllm~\cite{lu2024rtllm} have been shown to suffer from extensive contamination.
Recent studies even reveal that up to 40\% of LLM-generated code contains security vulnerabilities~\cite{he2024instruction}. 
Moreover, ongoing lawsuits alleging unauthorized reproduction of proprietary code by LLMs further underscore these risks~\cite{zhang2024right}. 
Together, these challenges raise a pressing question: \textbf{How can we intentionally mitigate the influence of problematic LLM memorization to achieve reliable hardware code generation?}

Existing efforts to mitigate problematic memorization in general LLMs have primarily focused on retraining and machine unlearning.
Retraining from scratch on sanitized datasets represents the most reliable way to achieve exact unlearning; yet, the prohibitive resource and time costs of training large-scale models make this approach largely impractical in real-world settings.
For instance, training the LLaMA 3.1 model (8B parameters)~\cite{llama3modelcard} from scratch requires approximately 1.46 million GPU hours on NVIDIA H100-80GB GPUs.
Machine unlearning has emerged as a more practical alternative~\cite{yao2024large, maini2024tofu, zhang2024negative, fan2024simplicity, wang2024llm, liu2025rethinking, rafailov2023direct}, which aims to selectively remove the influence of specific training data while preserving the utility of models.
This ensures that the resulting model behaves as if the removed data had never been included in training.
% \textcolor{red}{To achieve this, existing approaches adopt post-training strategies for efficient unlearning without retraining~\cite{yao2024large, maini2024tofu, zhang2024negative, fan2024simplicity, wang2024llm, liu2025rethinking, rafailov2023direct}, and have shown promise in natural language processing tasks, including harmful response removal, copyright erasure, and privacy protection. Recently, SALAD~\cite{wang2025salad} investigated the feasibility of applying existing unlearning techniques to hardware design, yet it suffered from substantial utility degradation and offered no domain-specific strategy. This gap highlights that the application of unlearning to hardware code generation, particularly to register-transfer level (RTL) code design, remains significantly underdeveloped.}
Specifically, existing efforts adopt post-training strategies for efficient unlearning without retraining~\cite{yao2024large, maini2024tofu, zhang2024negative, fan2024simplicity, wang2024llm, liu2025rethinking, rafailov2023direct}, and have shown promise in natural language processing tasks, including harmful response removal, copyright erasure, and privacy protection.
Yet, the application of unlearning to hardware code generation remains significantly underdeveloped.
Recently, only SALAD~\cite{wang2025salad} applied existing unlearning techniques to hardware design with limited domain adaptation, and therefore suffers from substantial utility degradation.
This gap highlights the need to develop domain-specific unlearning strategies to ensure reliable hardware code generation without compromising utility.

This paper proposes the first-of-its-kind domain-specific unlearning framework tailored to reliable hardware code generation, which is capable of eliminating problematic knowledge while maintaining reliable hardware code synthesis.
% Specifically, by recognizing that RTL codes require strict syntax, precise dependency modeling, and faithful execution semantics, our strategies essentially incorporate factors into unlearning processes to strike a balance between forgetting effectiveness and hardware code generation reliability.
Specifically, by recognizing that RTL codes require strict syntax, precise dependency modeling, and faithful execution semantics, our strategies incorporate these factors into unlearning processes to strike a balance between forgetting effectiveness and hardware code generation reliability.
%These domain ap enales allows LLMs to achieve significantly more efficient and stable unlearning compared to prior methods as shown by extensive experiments.
Our key contributions are as follows:

\begin{itemize}

   \item \textbf{Novel unlearning framework for reliable hardware code generation.} We pioneer the development of domain-specific unlearning frameworks to significantly improve the reliability of LLM-based hardware code generation with tailored strategies.
   
   %an RTL-specific mechanism that selectively excludes structural tokens from the forgetting process, ensuring that unlearning does not compromise the syntactic integrity of generated hardware code.

    \item \textbf{Syntax-preserving unlearning strategy.} We introduce an RTL-specific mechanism that selectively excludes structural tokens from the forgetting process, ensuring that unlearning does not compromise the syntactic integrity of generated hardware code.

    \item \textbf{Fine-grained floor-aware selective loss (FiFSL).} We develop a novel token-level objective that combines margin-based control with selective averaging, enabling precise and accelerated forgetting of undesired samples.

    \item \textbf{Scalable and robust unlearning for hardware code generation.} Comprehensive experiments show that our framework achieves up to 3$\times$ forget sets than traditional methods and requires only a single training epoch, while maintaining syntactic validity and functional reliability in RTL code generation, demonstrating state-of-the-art improvements in both efficiency, reliability, and utility.
\end{itemize}

\section{Background}
\label{sec:bg}

\subsection{Syntax and Semantics of Hardware Description Language}

A typical digital module is synchronous, consisting of two interacting components: \textbf{combinational logic}, which computes functions of the current inputs without memory, and \textbf{sequential logic}, which stores and updates states through registers or flip-flops at each clock edge. 
Register-Transfer Level (RTL) modeling captures this behavior by specifying data transfers between registers under clock control, together with the combinational logic that generates next-state values.
In Verilog, for instance, sequential logic is commonly expressed using \texttt{always @(posedge clk~or~negedge rst\_n)} blocks with non-blocking assignments (\texttt{<=}), while combinational logic is described with \texttt{assign} statements or \texttt{always @(*)} blocks using blocking assignments (\texttt{=}). 
Consequently, Verilog code for digital designs frequently relies on recurring syntax patterns, particularly \texttt{always} blocks that encode both combinational and sequential behavior.

\iffalse
A synchronous digital system is generally composed of two interacting parts: \textbf{combinational logic}, which computes pure functions of the current inputs without memory, and \textbf{sequential logic}, which uses registers or flip-flops to store state and update it on each clock edge.
Register--Transfer Level (RTL) modeling captures this behavior by describing data transfers between registers under clock control, along with the combinational logic that generates the next state values. For example, in Verilog, sequential logic is typically described using \texttt{always @(posedge clk~or~negedge rst\_n)} blocks with non-blocking assignments (\texttt{<=}), whereas combinational logic is implemented using \texttt{assign} statements or \texttt{always @(*)} blocks with blocking assignments (\texttt{=}).
As a result, when digital circuits are expressed in Verilog, a significant portion of the code often consists of these recurring syntax patterns---most notably the \texttt{always} blocks that represent both combinational and sequential behavior.
\fi

\subsection{Hardware Design with LLM}

\iffalse
    
Large Language Models (LLMs) are increasingly applied to hardware design automation, achieving notable success in Verilog generation, assertion creation, testbench synthesis, and EDA workflow optimization~\cite{thakur2023autochip, qiu2024autobench, wu2024chateda, liu2023chipnemo}.
Systems such as ChipNemo~\cite{liu2023chipnemo} and ChipGPT~\cite{chang2023chipgpt} demonstrate that fine-tuning and prompt engineering can substantially improve RTL generation quality, while advanced frameworks like HAVEN~\cite{yang2025haven}, CraftRTL~\cite{liu2024craftrtl}, and MAGE~\cite{zhao2024mage} extend capabilities through non-textual representations and multi-agent strategies. Benchmarks, including VerilogEval~\cite{liu2023verilogeval} and Rtllm~\cite{lu2024rtllm} confirm that LLMs can generate syntactically correct and functionally reliable RTL, highlighting their promise for accelerating design workflows.

\fi

%\textcolor{red}{Generative AI, especially Large Language Models (LLMs) are increasingly being applied to hardware development, achieving notable success in Verilog generation, assertion creation, testbench synthesis, and electronic design automation (EDA) workflow optimization~\cite{thakur2023autochip, qiu2024autobench, wu2024chateda, liu2023chipnemo,li2025analogfed, gao2025analoggenie, gaoanaloggenie, dong2023cktgnn, 11240792}.}

Generative AI is rapidly reshaping the landscape of hardware development~\cite{li2025analogfed, gao2025analoggenie, gaoanaloggenie, dong2023cktgnn, 11240792,poddar2025heart}. 
In particular, the integration of Large Language Models (LLMs) into digital hardware design has shown remarkable progress, demonstrating strong capability in Verilog generation, assertion synthesis, testbench creation, and optimization of electronic design automation (EDA) workflows~\cite{thakur2023autochip, qiu2024autobench, wu2024chateda, liu2023chipnemo}.
% \textcolor{red}{Beyond digital design, generative AI has also gained traction in analog hardware development, where models can automatically discover circuit topologies and support schematic exploration~\cite{li2025analogfed, gao2025analoggenie, gaoanaloggenie, dong2023cktgnn, 11240792}.}
% Building on this broader trend of AI-driven hardware design automation, 
Early efforts such as ChipNemo~\cite{liu2023chipnemo} and ChipGPT~\cite{chang2023chipgpt}, show that fine-tuning and prompt engineering can substantially improve RTL generation quality. 
More recent frameworks like HAVEN~\cite{yang2025haven}, CraftRTL~\cite{liu2024craftrtl}, and MAGE~\cite{zhao2024mage} extend these capabilities through non-textual representations and multi-agent strategies. 
Complementary benchmarks such as VerilogEval~\cite{liu2023verilogeval} and Rtllm~\cite{lu2024rtllm} further confirm that LLMs can produce syntactically correct and functionally reliable RTL, highlighting their promise to accelerate design workflows.

% \textcolor{red}{To Yiwen: make the below red part clearer, why they cannot provide strong robustness.}

Despite these advances, fundamental challenges remain in ensuring the safety, legality, and reliability of LLM-based hardware code generation.
Existing LLMs trained on massive public repositories risk propagating vulnerabilities and insecure coding practices through the memorization of buggy or outdated patterns.
To address these risks, security-focused techniques have emerged, such as SafeCoder~\cite{he2024instruction}, which augments training with curated secure-code datasets, and VersiCode~\cite{wu2024versicode}, a benchmark for evaluating robustness to Application Programming Interface (API) changes across releases.
Yet, these approaches fall short of providing comprehensive guarantees.
SafeCoder depends heavily on dataset quality and coverage, which cannot anticipate all insecure coding patterns or future vulnerabilities, leaving models susceptible to unseen exploits. 
VersiCode, by contrast, is limited to version-specific API robustness and does not address broader issues such as memorization of proprietary code, syntactic validity in hardware description languages, or broader security vulnerabilities.
These limitations highlight the need for novel approaches that can suppress unsafe behaviors while preserving the utility of LLMs for code generation.

%\textcolor{red}{While these efforts fall short of providing robust guarantees to address issues. }

\iffalse

Despite recent advances, critical challenges persist in safety, legality, and reliability.
LLMs trained on massive public repositories risk propagating vulnerabilities and insecure coding practices due to the memorization of buggy or outdated patterns.
To mitigate these risks, security-focused techniques such as SafeCoder~\cite{he2024instruction}, which augments training with curated secure-code datasets, and VersiCode~\cite{wu2024versicode}, a version-specific benchmark evaluating robustness to API changes across releases, have been proposed.
\textcolor{red}{While these efforts fall short of providing robust guarantees to address issues. }
Specifically, SafeCoder depends heavily on the quality and coverage of curated datasets, which cannot anticipate all possible insecure coding patterns or future vulnerabilities, leaving models susceptible to unseen exploits. 
VersiCode, on the other hand, is limited to version-specific API robustness and does not address deeper challenges such as memorization of proprietary code, syntactic validity in hardware description languages, or broader security vulnerabilities.
These limitations highlight the need for novel approaches—such as machine unlearning—that can suppress undesired or unsafe behaviors while preserving model utility in code generation.

\fi

\subsection{Machine Unlearning }

\iffalse

Machine unlearning—the process of eliminating the influence of specific data from a trained model without retraining from scratch—has emerged as an important research direction in recent years.
Within the domain of large language models (LLMs), a growing body of work in natural language processing (NLP) has explored unlearning techniques to address tasks such as privacy preservation, removal of copyrighted material, and suppression of harmful content.
Representative methods include Gradient Ascent (GA)~\cite{yao2024large}, which attempts to “reverse” prior learning by performing ascent on the next-token prediction loss of the forget set; however, GA often suffers from instability and collateral drift on the retain distribution due to its indiscriminate and unbounded optimization signal.
To mitigate these issues, Negative Preference Optimization (NPO)~\cite{zhang2024negative} introduces an alignment-inspired loss function that embeds a smooth, bounded penalty encouraging divergence from forget-set behavior while maintaining model stability.
\textcolor{red}{To Yiwen: more key details on SimNPO.}
More recently, SimNPO~\cite{fan2024simplicity} was proposed as an improved variant of NPO that eliminates the reliance on a teacher model and instead uses a reference-free, length-normalized objective for unlearning.
Therefore, it mitigates the reference bias problem in NPO, which can lead to uneven allocation of unlearning power and unstable gradient smoothing, while achieving more balanced forgetting and stronger utility preservation.

\fi

Machine unlearning, the process of eliminating the influence of specific data from a trained model without retraining from scratch, has recently emerged as a critical research field.
In the context of LLMs, unlearning has been increasingly explored to address applications such as privacy preservation, removal of copyrighted material, and suppression of harmful content~\cite{liu2025rethinking}.
Early methods, such as Gradient Ascent (GA)~\cite{yao2024large}, attempt to ``reverse'' prior learning by performing ascent on the next-token prediction loss of the forget set, but they often suffer from instability and collateral drift on the retain distribution due to the indiscriminate and unbounded optimization signal.
To tackle these issues, Negative Preference Optimization (NPO)~\cite{zhang2024negative} introduces an alignment-inspired loss function embedded with a smooth bounded penalty to promote divergence from forget-set behavior while maintaining stability.
More recently, SimNPO~\cite{fan2024simplicity} refines this approach by removing the reliance on a teacher model and adopting a reference-free, length-normalized objective, thereby mitigating reference bias and enabling more balanced forgetting with stronger utility preservation.

% Therefore, it mitigates the reference bias problem in NPO, which can lead to uneven allocation of unlearning power and unstable gradient smoothing, while achieving more balanced forgetting and stronger utility preservation.

While these advances have demonstrated the feasibility of unlearning in natural language processing (NLP), their applicability to hardware code generation remains largely unexplored. 
Unlike natural language, RTL code exhibits strict syntactic and structural constraints, where even minor inconsistencies can render outputs unsynthesizable. 
As a result, NLP-oriented unlearning methods cannot be directly transferred to hardware description languages--as shown by a recent work SALAD~\cite{wang2025salad}, ignoring domain-specific syntax significantly undermines the validity of code generation.
These limitations highlight the need for domain-specific unlearning strategies that not only suppress undesired behaviors but also preserve the correctness and reliability of RTL code generation for hardware design.

\iffalse

While these advances have demonstrated the feasibility and utility of unlearning in NLP, their applicability to \textcolor{red}{code generation}, especially hardware description language, remains largely unexplored. 
Unlike natural language, RTL code exhibits strict syntactic and structural constraints, where even minor inconsistencies can render outputs unsynthesizable. Thus, directly transferring NLP-oriented unlearning methods is insufficient. 
These limitations underscore the need for novel unlearning strategies to ensure that undesired behaviors are effectively suppressed while maintaining the correctness and reliability of RTL code generation for hardware design.

\fi

\section{Methodology}
\label{sec:method}

% \textcolor{red}{To Yiwen: An overview of your method at the beginning of this section is good.}
% \textcolor{red}{To Yiwen: not doing well here.}

\subsection{Framework Overview}
\label{sec:overview}

\iffalse
As shown in Fig.~\ref{fig:overview}, an LLM trained on massive datasets may inadvertently memorize undesirable knowledge, such as proprietary IP and unsafe coding patterns, which can lead to illegal or unreliable RTL outputs during hardware generation.
To mitigate this issue.
we propose a domain-specific unlearning framework that leverages harmful data as a forget set, enabling the removal of problematic knowledge while preserving code utility.
Our approach combines two techniques: 
\textbf{(1) syntax-preserving masking}, which protects reserved keywords and code spans to maintain compilability, and 
\textbf{(2) fine-grained floor-aware selective loss (FiFSL)}, which applies margin-based forgetting pressure while gating out already forgotten samples to focus updates on the hardest cases. 
Together, these strategies enable efficient, stable, and utility-preserving unlearning, yielding LLMs that generate reliable hardware code.
\fi

\begin{figure}[!t]
 % \vskip -12pt
  \centering
\includegraphics[width=1.0\columnwidth]{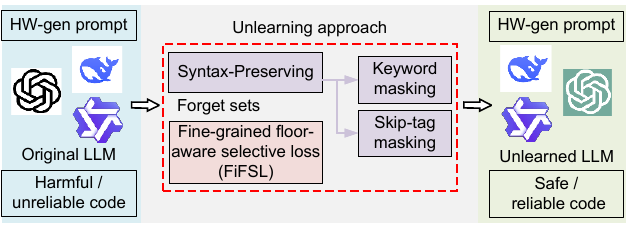}
 \vskip -6pt
\caption{Overview of our proposed domain-specific LLM unlearning framework for reliable hardware code generation (HW-gen).}
\label{fig:overview}
 \vskip -12pt
\end{figure}

% As shown in Fig.~\ref{fig:overview}, an LLM trained on massive datasets may inadvertently memorize undesirable knowledge, such as proprietary IP and unsafe coding patterns, which can lead to illegal or unreliable RTL outputs during hardware generation.
% To mitigate this issue.
We propose a domain-specific unlearning framework for reliable LLM-based hardware code generation.
Fig.~\ref{fig:overview} shows its overview, which leverages harmful data as a forget set, enabling the removal of problematic knowledge while preserving code utility.
Our approach combines two major techniques: 
\textbf{(1) syntax-preserving masking}, which protects reserved keywords and code spans to maintain compilability, and 
\textbf{(2) fine-grained floor-aware selective loss (FiFSL)}, which applies margin-based forgetting pressure while gating out already forgotten samples to focus updates on the hardest cases. 
Together, these strategies enable efficient, stable, and utility-preserving unlearning, yielding LLMs that generate reliable hardware codes.

\subsection{Problem Definition}
\label{sec:problem_defin}

We consider the problem of machine unlearning in the context of LLMs for RTL code generation.
Let $\pi_\theta$ denote a pretrained LLM parameterized by $\theta$, which generates RTL code $y$ given an instruction or design intent $x$, i.e., $\pi_\theta(y|x)$. 
In practice, not all training data should be retained indefinitely. 
Certain samples may correspond to privacy-sensitive hardware designs, erroneous code snippets, or legally restricted RTL modules~\cite{liu2023verilogeval}. Such cases can arise due to incomplete data cleaning, outdated versioned code, etc.
We denote the set of data to be forgotten as $D_f = {(x_f, y_f)}$.
The objective of unlearning is to update the model into $\pi_\theta^*$ such that its dependence on $D_f$ is minimized.
Specifically, given a prompt $x_f$, the outputs of $\pi_\theta^*$ should diverge from the undesired targets $y_f$.
Meanwhile, the model must preserve its utility on the remaining distribution.
We denote the validation set by $D_v = {(x_v,y_v)}$, which reflects the task-relevant code generation capability to be preserved.
Formally, the unlearning goal can be expressed as:
\begin{equation}
    \begin{aligned}
        \text{Find}\ \pi_\theta^\ast \quad \text{s.t.}\ \
        &\pi_\theta^\ast(y_f \mid x_f) \ \text{is minimized},  &\forall (x_f,y_f) \in D_f, \\
        &\pi_\theta^\ast(y_v \mid x_v) \ \approx \ \pi_\theta(y_v \mid x_v), &\forall (x_v,y_v) \in D_v.
    \end{aligned}
    \label{eq:unlearning}
\end{equation}
This formulation captures the dual goal of unlearning: removing the influence of undesired data while retaining the model’s competence in hardware code generation.
To address this dual goal in practice, our methodology consists of two key components: (i) syntax-preserving unlearning that ensures syntax structural correctness is not disrupted, and (ii) a fine-grained floor-aware selective loss function that accelerates and strengthens the forgetting process.

%\textcolor{red}{To Yiwen: Divide the paragraph below into two parts: background and technical.}

\subsection{Proposed Unlearning Framework}
\subsubsection{Syntax-Preserving Unlearning}
\label{sec:mask}
\begin{figure}[!t]
 % \vskip -5pt
  \centering
\includegraphics[width=1.0\columnwidth]{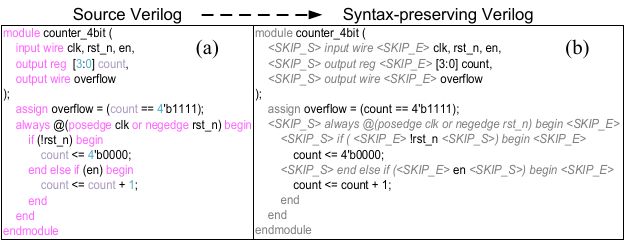}
 \vskip -6pt
\caption{Syntax-preserving unlearning: source Verilog (a) vs. syntax-preserving Verilog (b). The preserved syntax is masked in gray.
}
\label{fig:Syntax-Preserving}
 \vskip -12pt
\end{figure}

Unlike free-form natural languages, RTL codes follow a finite and well-defined grammar.
Their tokens, such as reserved keywords, operators, delimiters, and numeric literals, recur systematically to capture combinational and sequential behaviors.
While exact idioms differ across cases (e.g., continuous assignments or sensitivity lists for combinational logic vs. clocked processes for sequential logic), such tokens represent universal syntactic backbones rather than task-specific semantics.
% Thus, unlearning strategies that indiscriminately suppress these high-frequency syntax tokens are harmful: they undermine grammaticality and compromise the compilability and synthesizability of RTL programs.
Thus, existing unlearning strategies~\cite{wang2025salad} that indiscriminately suppress these high-frequency syntax tokens are harmful: they undermine grammaticality and compromise the compilability and synthesizability of RTL programs (i.e., treating every token in the toy example Verilog code of a 4-bit counter (Fig.~\ref{fig:Syntax-Preserving}(a)) as removable).
Effective unlearning must be syntax-preserving, targeting semantic content while enforcing grammar constraints to maintain well-formed hardware description languages (HDLs).

To achieve this, we propose a \textbf{syntax-preserving unlearning strategy that selectively protects structural elements while forgetting only task-specific tokens}.
As shown in Fig.~\ref{fig:Syntax-Preserving}(b), our approach integrates two complementary mechanisms to preserve structural correctness, grounded in the IEEE hardware code standard~\cite{10458102, 1620780, palnitkar2003verilog}, and excludes them from the loss function.
First, keyword masking is applied to a curated set of Verilog reserved words such as \textit{module}, \textit{assign}, \textit{wire}, etc, ensuring that these essential tokens are not modified by forgetting operations.
 % (the blue color in Fig.~\ref{fig:Syntax-Preserving}(b))
Second, skip-tag masking introduces special tokens \verb|<SKIP_S>| and \verb|<SKIP_E>| to mark spans of texts that should be ignored during loss computation, and all tokens appearing between these tags are excluded from the unlearning objective.
% (the gray parts in Fig.~\ref{fig:Syntax-Preserving}(b)).
To unify both operations, we assign each token $x_t$ in an input sequence $\mathbf{x} = (x_1, \ldots, x_T)$ with a binary mask value $m_t$, where $m_t = 0$ if $x_t$ falls within a skip-tag span or belongs to the reserved keyword set $\mathcal{K}$, and $m_t = 1$ otherwise. 
The loss for syntax-preserving unlearning is thus computed directly as
\begin{equation}
    \mathcal{L}_{\text{syntax}} = \sum\nolimits_{t=1}^T m_t \, \ell\!\big(\pi_\theta(x_t \mid \mathbf{x}_{<t}), y_t)
    \label{eq:element_loss}
\end{equation}
with $\, \ell\!\big (\cdot)$ denoting the token-level cross-entropy.
By aligning the masking rule with the loss function, the model forgets undesired content while maintaining the universal syntactic backbone required for compilable RTL code.
It is important to note that from the computational perspective, the proposed masking strategy introduces negligible overheads, since it only requires constructing a token-level binary mask during preprocessing and applying it element-wise to the loss function. 
This makes syntax-preserving unlearning highly efficient and scalable, enabling its deployment in large-scale Verilog LLMs without incurring noticeable computational or memory costs.

\subsubsection{FiFSL: Fine-grained Floor-aware Selective Loss}
\label{sec:loss}
To further enhance the unlearning process, we introduce a \textbf{fine-grained floor-aware selective loss (FiFSL)} that operates at the token level and combines margin-based control with a selective averaging mechanism.
The key idea is to impose a lower bound on the loss to prevent excessive forgetting, while gradients are computed only for unmasked tokens and unforgotten samples.

\iffalse

To further enhance the unlearning process, we introduce a \textbf{fine-grained floor-aware selective loss (FiFSL)} that operates at the token level and combines margin-based control with a selective averaging mechanism.
In general, we impose a lower bound on the loss to prevent excessive forgetting, while gradients are computed only for unmasked tokens and unforgotten samples.
%\textcolor{red}{To Yiwen: lack of one or two sentences to show the main principle/idea of this technique at a high level.}

\fi

\textbf{Forward:}
In the forward pass, for each sample $i$ in a batch of size $b$, let $\mathcal{V}_i$ denote the set of all valid (unmasked) tokens after the syntax-preserving masking in Sec.~\ref {sec:mask}.
The token-normalized negative log-likelihood for sample $i$ is defined as
\begin{equation}
\begin{aligned}
            L_{i} &= 1/{|\mathcal{V}_{i}|}\cdot \mathcal{L}_{\text{syntax}}  \\
           % &= \frac{1}{|\mathcal{V}_{i}|} 
   % \sum_{t \in \mathcal{V}_{i}} m_t
    %\ell\!\big(\pi_\theta(x_{t} \mid \mathbf{x}_{<t}), y_{t}\big).
\end{aligned}
    \label{eq:loss}
\end{equation}
To impose forgetting, we shift this loss by a margin parameter $\gamma$ and apply a smooth negative-preference mapping, yielding
\begin{equation}
\begin{aligned}
    \phi_{i} &= {2}/{\beta} \, \mathrm{softplus}\!\left(-\beta \bigl(L_{i} - \gamma \bigr)\right),
    %\\
   % &= \frac{2}{\beta} \, \mathrm{softplus}\!\left(-\beta \bigl( \frac{1}{|\mathcal{V}_{i}|} 
   % \sum_{t \in \mathcal{V}_{i}} m_t
   % \ell\!\big(\pi_\theta(x_{t} \mid \mathbf{x}_{<t}), y_{t}\big) - \gamma \bigr)\right),
\end{aligned}
\label{eq:phi}
\end{equation}
where $\beta > 0$ controls the curvature and saturation rate (larger $\beta$ gives a sharper transition), and $\mathrm{softplus}(z)=\log (1 + e^z)$ is used to provide a smooth, bounded mapping.
This construction exerts strong forgetting pressure when a sample is under-forgotten ($L_{i} < \gamma$), and gradually saturates once the margin is exceeded, thus preventing runaway gradients.
To further prevent excessive forgetting and suppress outlier effects, we introduce a hard floor $L_\text{min}$ and average only over active samples whose penalties exceed this threshold ($\phi_i > L_{\min}$), contributing both to the loss and to the gradient.
Let $\alpha_i = \mathbf{1}\{\phi_{i} > L_{\min}\}$ and $N_{\text{act}} = \sum_i \alpha_i$, the mini-batch objective loss is expressed as
\begin{equation}
    \begin{aligned}
        \mathcal{L}_{\mathrm{FiFSL}}^{\text{batch}} = 
       {1}/{N_{\mathrm{act}}} 
        \sum\nolimits_{i} a_{i} \, \phi_{i} . 
    %     =
    %     \frac{1}{N_{\mathrm{act}}} 
    %     \sum_{i} a_{i} \cdot \frac{2}{\beta} \, \mathrm{softplus}\!\left(-\beta \bigl( \frac{1}{|\mathcal{V}_{t}|} 
    % \sum_{t \in \mathcal{V}_{t}}
    % \ell\!\big(\pi_\theta(x_{t} \mid \mathbf{x}_{<t}), y_{t}\big) - \gamma \bigr)\right).
    \end{aligned}
    \label{eq:fifsl_batch}
\end{equation}
This is algebraically equivalent to first clamping $\phi_i$ by $L_{\min}$, masking inactive samples, and then normalizing by the number of active ones. 
As a result, samples with $\phi_i \leq L_{\min}$ contribute no gradient and are effectively excluded from further updates.
At the dataset level, FiFSL can be expressed as the expectation over the forget set $D_f$:
\begin{equation}
\begin{aligned}
    \mathcal{L}_{\mathrm{FiFSL}}(\theta)
&= \mathbb{E}{(x,y)\in D_f}\Big[\alpha(x,y) \cdot \phi\big(\pi_\theta, x, y\big)\Big], \\ &\alpha(x,y)  = \mathbf{1}\{\phi(\pi_\theta, x, y) > L_{\min}\}.
\end{aligned}
\end{equation}

% \textcolor{blue}{Figures per\_sample, per\_batch}

\textbf{Backpropagation:}
By establishing the forward objective, we now analyze the backward signal to show how our loss enforces forgetting stably and selectively.
Recall Eqs.~\eqref{eq:loss} and ~\eqref{eq:phi}.
Using $\mathrm{softplus}'(z) = \sigma (z)$ with the $\sigma$ as logistic sigmoid, differentiating $\phi_i$ with respect to the per-sample loss yields
\begin{equation}
    {\partial \phi_i}/{\partial L_i}
= -2 \sigma\!\left(-\beta \big(L_i - \gamma\big)\right).
\label{eq:diff_phi}
\end{equation}
Thus, the sensitivity lies strictly between $-2$ and $0$: when $ L_{i} < \gamma$, the term $-\beta \big(L_i - \gamma\big)$ is positive, $\sigma(\cdot) \approx 1$, and the gradient magnitude approaches two, producing a strong negative push that increases $L_i$ and drives forgetting.
Consequently, when $ L_{i} > \gamma$, the sigmoid gradually reduces the gradient, leading to milder updates and preventing overly large gradients that could cause excessive forgetting.
At the token level, the derivative of $ L$ with respect to a logit at position $t$ is $\frac{1}{|\mathcal{V}_i|} \, m_{i,t} \big( \mathbf{p}_{i,t} - \mathbf{e}(y_{i,t}) \big)$.
% \begin{equation}
%     \frac{1}{|\mathcal{V}_i|} \, m_{i,t} \big( \mathbf{p}_{i,t} - \mathbf{e}(y_{i,t}) \big),
% \end{equation}
Here, $m_{i,t}$ is the syntax mask, $\mathbf{p}_{i,t}$ the softmax vector, and 
$\mathbf{e}(y_{i,t})$ the one-hot target.
The normalization by $|\mathcal{V}_i|$ ensures that gradients are comparable across sequences of different lengths, while the mask guarantees that reserved keywords or skipped tokens never receive updates.
Combining these results, the batch-level gradient becomes
\begin{equation}
\frac{\partial \mathcal{L}^{\text{batch}}_{\text{FiFSL}}}{\partial \mathrm{logit}_{i,t}} = \frac{\alpha_i}{N_{\text{act}}}
\left(-2\sigma\!\left(-\beta \left(L_{i} - \gamma\right)\right)\right)
\frac{1}{|\mathcal{V}_i|}
\, m_{i,t}\big(\mathbf{p}_{i,t} - \mathbf{e}(y_{i,t})\big).
\end{equation}
This expression shows that FiFSL rescales the standard cross-entropy gradient by a bounded negative factor and suppresses it entirely when either the token is masked or the sample is already forgotten $(\alpha_i = 0)$.

The implications are twofold. First, FiFSL enforces forgetting with control: active samples below the margin receive strong forgetting pressure, but this pressure decays smoothly once the margin is passed, stabilizing optimization. Second, FiFSL avoids unnecessary updates by gating entire samples once they have been sufficiently forgotten, unlike conventional unlearning techniques~\cite{yao2024large, zhang2024negative, fan2024simplicity}, which continue to update all samples uniformly. 
In practice, as training proceeds, the number of active samples decreases, and optimization effort concentrates on the hard-to-forget tails of the distribution. This selective mechanism not only prevents collateral degradation of syntax tokens but also reduces the number of epochs required to achieve effective unlearning compared to conventional methods.

\subsubsection{Summary of the Proposed Strategy}
In summary, our framework combines \textit{syntax-preserving unlearning}, 
which safeguards the structural correctness of RTL code, 
with \textit{FiFSL}, a margin-based selective forgetting loss that ensures stable 
and efficient removal of undesired knowledge. 
This joint design enables effective forgetting while preserving syntactic and functional validity.
% The proposed procedure is summarized in Algorithm~\ref{alg:unlearning}.
The process is presented in Algorithm~\ref{alg:unlearning}.
\begin{algorithm}[!t]
\small
\caption{Our domain-specific unlearning method.}
\label{alg:unlearning}
\KwIn{Forget set $\mathcal{D}_f$, model $\pi_\theta$, reserved keywords $K$, skip-tags}
\ForEach{batch $(x,y) \in \mathcal{D}_f$}{
    % --- Collator: Syntax-preserving masking ---
    \tcp{Syntax-preserving}
    \ForEach{token $x_t$ in $x$ }{  
        \eIf{$x_t \in K$ or inside skip-tags }{
            $m_t \gets 0$, \; $labels[t] \gets -100$ \tcp{ignored in loss}
        }{
            $m_t \gets 1$
        }
    }

    % --- Syntax loss ---
    $L_{\text{syntax}} \gets \sum_t m_t \cdot \ell\! \ (\pi_\theta(x_t|x_{<t}), y_t)$\;

    % --- FiFSL selective forgetting ---
    \tcp{FiFSL}
    \ForEach{sample $i$ }{   
        $L_i \gets \text{normalized}(L_{\text{syntax}}(i))$\;
        $\phi_i \gets \tfrac{2}{\beta}\,\text{softplus}(-\beta(L_i-\gamma))$\;
        $\alpha_i \gets 1$ if $\phi_i > L_{\min}$ else $0$\;
    }

    % --- Batch update ---
    $L_{\text{batch}} \gets \tfrac{\sum_i \alpha_i \cdot \phi_i}{\sum_i \alpha_i}$\;
    $\theta \gets \theta - \eta \nabla_\theta L_{\text{batch}}$\;
}

\end{algorithm}
\section{Experiments}
\label{sec:experiment}

\begin{table}[!t]
% \vskip -9pt
\centering
\setlength{\tabcolsep}{5.5pt}
\caption{ Performance of benchmark models after fine-tuning.}
\vskip -6pt
\begin{tabular}{|c|cc|cccc|}
\hline
\multirow{2}{*}{Model} & \multicolumn{2}{c|}{\textbf{Forget quality}} & \multicolumn{4}{c|}{\textbf{Model utility}} \\ \cline{2-7}
 & PrivLeak & MinK++ & Loss & Pass@1 & BLEU & chrF \\ \hline
Llama-7B & 0.98     & 0.80   & 0.30 & 49\%   & 50.7 & 59.1 \\ \hline
Llama-8B & 0.96     & 0.97   & 0.41 & 53\%   & 40.1 & 56.2 \\ \hline
DeepSeek-7B & 0.97     & 0.83   & 0.28 & 55\%   & 48.5 & 57.9 \\ \hline
Qwen-7B & 0.87     & 0.96   & 0.25 & 57\%   & 50 & 57.7 \\ \hline

\end{tabular}
\label{tab:baseline}
\vskip -12pt
\end{table}

\begin{figure*}[!t]
 % \vskip -12pt
  \centering
\includegraphics[width=0.95\linewidth]{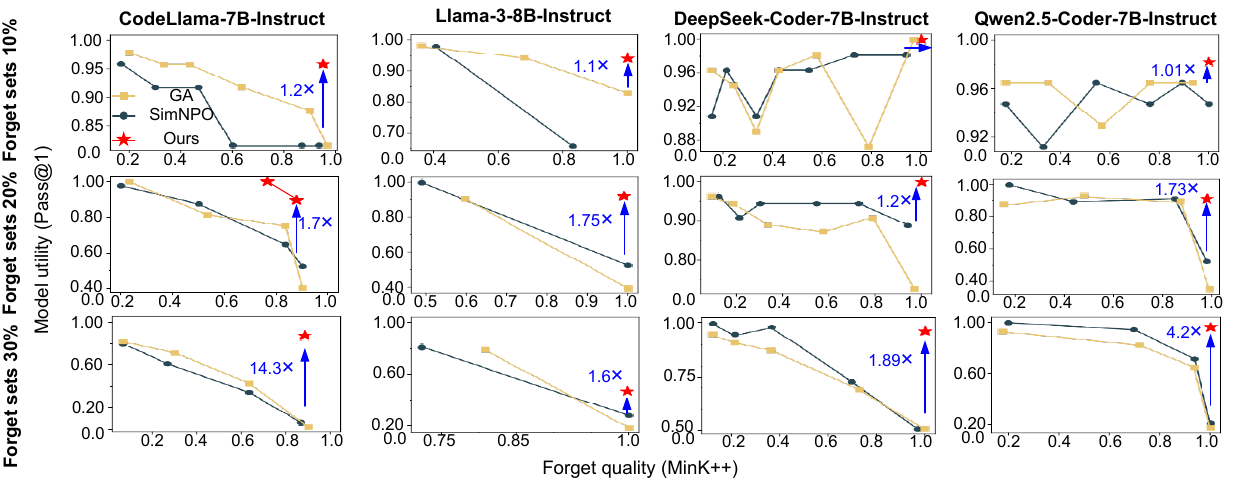}
 \vskip -12pt
\caption{Comparisons of forgetting quality and model utility across unlearning methods, varying forget-set sizes, and different LLMs. Ideal unlearning target: $(1.0, 1.0)$, perfect forgetting with no utility loss. Our method achieves comparable or better trade-offs in only 1 epoch, while baselines require 4$\sim$7 epochs.
}
\label{fig:Results}
 \vskip -12pt
\end{figure*}

\subsection{Experimental Methodology}

\subsubsection{Evaluation Metrics}
Evaluations are conducted along two dimensions: (i) \textbf{forget quality} and (ii) \textbf{model utility}.
These metrics are widely adopted in the LLM unlearning domain and are adapted here to the context of hardware code generation.

\textbf{Forget quality.}
We employ two metrics to quantify the degree of forgetting:
(1) \textit{PrivLeak}~\cite{nasr2023scalable}, which quantifies the probability that the model regenerates HDL snippets or modules from the forget set. This corresponds to verbatim or near-verbatim leakage of RTL blocks (e.g., always blocks, state machines, or functional units). 
\textit{Lower values indicate stronger forgetting and reduced IP leakage risk}.
(2) \textit{MinK++}~\cite{zhang2024min}, which evaluates the average log-likelihood of the lowest-$k\%$ gold tokens in generated HDL codes. 
By calibrating against token distributions, it is sensitive to residual memorization of hardware coding patterns (e.g., signal naming conventions, pipeline templates) even when direct leakage does not occur. 
\textit{Lower scores suggest weaker memorization}.
For reference, untrained models typically yield scores around $0.5$~\cite{zhang2024min}.
\textbf{Model utility.}
To ensure that forgetting does not degrade downstream performance, we evaluate model utility on hardware code generation tasks using:
(1) \textit{Cross-entropy loss} on held-out HDL test sets, measuring token-level predictive quality. 
(2) \textit{Pass@1 accuracy} on functional hardware code generation benchmarks, assessing whether the generated Verilog/HDL compiles and passes simulation testbenches.
(3) \textit{BLEU} and \textit{chrF}, which capture syntactic and stylistic fidelity.
BLEU measures $n$-gram overlap with reference RTL implementations, while chrF evaluates character-level similarity, useful for ensuring correct use of hardware-specific syntax.
% \textcolor{red}{BLEU measures n-gram overlap with reference RTL implementations.
% For example, matching opcode sequences such as ``\texttt{always @ (posedge clk)}” across candidate and reference code.
% chrF, in contrast, evaluates character-level similarity, useful for ensuring correct use of hardware-specific syntax (e.g., \texttt{rst\_n/rest\_n}).}

\subsubsection{Benchmarks and Experiment Setup}
We evaluate our approach on four LLMs fine-tuned for RTL code generation: Llama-3-8B-Instruct~\cite{llama3modelcard}, CodeLlama-7B-Instruct~\cite{roziere2023code}, DeepSeek-Coder-7B-Instruct~\cite{deepseek-coder}, and Qwen2.5-Coder-7B-Instruct~\cite{hui2024qwen2}. 
Among them, Llama-3-8B-Instruct is a general-purpose model, while others are explicitly optimized for code understanding and generation. 
As the fine-tuned corpus, we adopt the open-source RTLCoder dataset~\cite{liu2024rtlcoder}, which provides instruction-response pairs comprising natural language design specifications and their corresponding RTL code implementations.
Each model is fine-tuned on 1,000 samples for six epochs using the Adam optimizer with a learning rate of $1\times10^{-5}$, a maximum sequence length of 2,048 tokens, and a batch size of 2 on 4$\times$A6000 GPUs. 
For inference, we set the sampling temperature to 0.5 and top-$p$ to 0.9.
Table~\ref{tab:baseline} reports the performance of these benchmark models after fine-tuning. 
All models exhibit strong memorization tendencies (PrivLeak and MinK++ $>0.8$), which provides a baseline reference for subsequent unlearning evaluations.

Since there are no publicly available domain-specific datasets for proprietary or contaminated RTL designs, we follow prior works~\cite{yao2024large, maini2024tofu, zhang2024negative, fan2024simplicity, wang2024llm, liu2025rethinking, rafailov2023direct, wang2025salad} to emulate realistic unlearning scenarios by subsampling 100, 200, and 300 examples from the training corpus (i.e., about 10\%, 20\%, and 30\%).
% To emulate realistic unlearning scenarios, we construct forget sets by subsampling 100, 200, and 300 examples from the training corpus (i.e., about 10\%, 20\%, and 30\%).
\textit{These forget sets emulate practical cases where LLMs may memorize proprietary IP, contaminated benchmarks, or unsafe coding patterns that must be forgotten}.
The downstream utility is assessed on the held-out RTLCoder test set, consistent with the benchmark evaluation protocol.

{\renewcommand{\arraystretch}{0.9}
\begin{table*}[!t]
\centering
 % \vskip -5pt
\caption{A Comprehensive Comparison of Model Performance After Unlearning on Different Sizes of Forget Sets.}
\vskip -5pt
\scriptsize
\small
\resizebox{\textwidth}{!}{
\begin{tabular}{l c l c c c | c c c c | c c c c}
\toprule
\multirow{2}{*}{Model} & \multirow{2}{*}{Forget sets} & \multirow{2}{*}{Method} & \multirow{2}{*}{Epoch $\downarrow$} & \multirow{2}{*}{PrivLeak $\downarrow$} & \multirow{2}{*}{MinK++ $\downarrow$}
& \multicolumn{4}{c|}{\textbf{Validation}} & \multicolumn{4}{c}{\textbf{Generalization Ability}} \\
\cmidrule(lr){7-10}\cmidrule(lr){11-14}
& & & & & & Loss $\downarrow$ & Pass@1 $\uparrow$ & BLEU $\uparrow$ & chrF $\uparrow$
& Loss $\downarrow$ & Pass@1 $\uparrow$ & BLEU $\uparrow$ & chrF $\uparrow$ \\
\midrule
%================ Llama-7B ==================
\multirow{10}{*}{Llama-7B}
& --- & --- & --- & 0.98 & 0.80 & 0.30 & 49\% & 50.7 & 59.1 & 0.35 & 73\% & 34.5 & 59.6 \\
\cmidrule(lr){2-14}
& \multirow{3}{*}{10\%} & GA     & 6 & 0.75 & 0.52 & 0.46 & 40\% & 40.5 & 50.3 & 0.70 & 68\% & 42.5 & 61.6 \\
&                      & SimNPO & 6 & 0.75 & 0.51 & 0.46 & 40\% & 41.6 & 51.2 & 0.70 & 62\% & 40.6 & 61.1 \\ 
&                      & \cellcolor{gray!20}Ours   & \cellcolor{gray!20}1 & \cellcolor{gray!20}0.70 & \cellcolor{gray!20}0.53 & \cellcolor{gray!20}0.30 & \cellcolor{gray!20}47\% (-2\%) & \cellcolor{gray!20}50.6 & \cellcolor{gray!20}57.9 & \cellcolor{gray!20}0.34 & \cellcolor{gray!20}73\% & \cellcolor{gray!20}46.0 & \cellcolor{gray!20}67.0 \\
\cmidrule(lr){2-14}
& \multirow{3}{*}{20\%} & GA     & 4 & 0.82 & 0.53 & 0.48 & 26\% & 40.5 & 48.8 & 0.67 & 53\% & 31.5 & 57.0 \\
&                      & SimNPO & 4 & 0.82 & 0.53 & 0.50 & 20\% & 40.5 & 48.0 & 0.73 & 48\% & 30.0 & 55.8 \\
&                      & \cellcolor{gray!20}Ours   & \cellcolor{gray!20}2 & \cellcolor{gray!20}0.65 & \cellcolor{gray!20}0.54 & \cellcolor{gray!20}0.33 & \cellcolor{gray!20}45\% (-4\%) & \cellcolor{gray!20}48.8 & \cellcolor{gray!20}56.4 & \cellcolor{gray!20}0.43 & \cellcolor{gray!20}69\% & \cellcolor{gray!20}35.3 & \cellcolor{gray!20}59.1 \\
\cmidrule(lr){2-14}
& \multirow{3}{*}{30\%} & GA     & 4 & 0.73 & 0.54 & 1.00 & 3\% &  13.2 & 28.7 & 1.70 & 5\% & 8.2 & 26.6 \\
&                      & SimNPO & 4 & 0.72 & 0.53 & 1.00 & 1\% &  11.7 & 26.9 & 1.70 & 4\% & 8.0 & 24.7 \\
&                      & \cellcolor{gray!20}Ours   & \cellcolor{gray!20}1 & \cellcolor{gray!20}0.65 & \cellcolor{gray!20}0.53 & \cellcolor{gray!20}0.32 & \cellcolor{gray!20}44\% (-5\%) & \cellcolor{gray!20}49.7 & \cellcolor{gray!20}56.8 & \cellcolor{gray!20}0.40 & \cellcolor{gray!20}72\% & \cellcolor{gray!20}37.9 & \cellcolor{gray!20}60.0 \\
\midrule
%================ Llama-8B  ==================
\multirow{10}{*}{Llama-8B }
& --- & --- & --- & 0.96 & 0.97 & 0.41 & 53\% & 40.1 & 56.2 & 0.50 & 70\% & 27.1 & 56.0 \\
\cmidrule(lr){2-14}
& \multirow{3}{*}{10\%} & GA     & 2 & 0.66 & 0.58 & 0.50 & 35\% & 35.0 & 50.2 & 0.70 & 45\% & 21.7 & 47.0 \\
&                      & SimNPO & 3 & 0.61 & 0.50 & 0.51 & 44\% & 37.8 & 51.1 & 0.80 & 50\% & 21.7 & 47 \\
&                      & \cellcolor{gray!20}Ours   & \cellcolor{gray!20}1 & \cellcolor{gray!20}0.68 & \cellcolor{gray!20}0.49 & \cellcolor{gray!20}0.31 & \cellcolor{gray!20}50\% (-3\%) & \cellcolor{gray!20}47.0 & \cellcolor{gray!20}55.6 & \cellcolor{gray!20}0.33 & \cellcolor{gray!20}73\% & \cellcolor{gray!20}44.2 & \cellcolor{gray!20}62.4 \\
\cmidrule(lr){2-14}
& \multirow{3}{*}{20\%} & GA     & 2 & 0.67 & 0.50 & 0.66 & 28\% & 30.3 & 44.2 & 1.10 & 30\% & 17.0 & 42.1 \\
&                      & SimNPO & 2 & 0.69 & 0.50 & 0.59 & 21\% & 24.7 & 44.9 & 0.97 & 25\% & 12.5 & 37.0 \\
&                      & \cellcolor{gray!20}Ours   & \cellcolor{gray!20}1 & \cellcolor{gray!20}0.64 & \cellcolor{gray!20}0.51 & \cellcolor{gray!20}0.34 & \cellcolor{gray!20}49\% (-4\%) & \cellcolor{gray!20}44.0 & \cellcolor{gray!20}51.8 & \cellcolor{gray!20}0.40 & \cellcolor{gray!20}71\% & \cellcolor{gray!20}41.9 & \cellcolor{gray!20}58.3 \\
\cmidrule(lr){2-14}
& \multirow{3}{*}{30\%} & GA     & 2 & 0.65 & 0.50 & 0.77 &  15\% &  16.8 & 36.9 & 1.20 & 24\% & 8.0 & 27.6 \\
&                      & SimNPO & 2 & 0.65 & 0.50 & 0.78 &  10\% & 16.0 & 37.3 & 1.28 &  18\% & 7.5 & 27.5 \\
&                      & Ours   & 1 & 0.60 & 0.51 & 0.37 & 24\% & 35.0 & 50.0 & 0.53 & 34\% & 21.8 & 50.5 \\
\midrule
%================ DeepSeek ==================
\multirow{10}{*}{DeepSeek-7B}
& --- & --- & --- & 0.97 & 0.83 & 0.28 & 55\% & 48.5 & 57.9 & 0.30 & 80\% & 37.3 & 59.8 \\
\cmidrule(lr){2-14}
& \multirow{3}{*}{10\%} & \cellcolor{gray!20}GA     & \cellcolor{gray!20}7 & \cellcolor{gray!20}0.77 & \cellcolor{gray!20}0.52 &\cellcolor{gray!20}0.35 & \cellcolor{gray!20}54\% & \cellcolor{gray!20}44.2 & \cellcolor{gray!20}53.3 & \cellcolor{gray!20}0.42 & \cellcolor{gray!20}81\% & \cellcolor{gray!20}44.4 & \cellcolor{gray!20}61.9 \\
&                      & \cellcolor{gray!20}SimNPO & \cellcolor{gray!20}7 & \cellcolor{gray!20}0.75 & \cellcolor{gray!20}0.50 & \cellcolor{gray!20}0.37 & \cellcolor{gray!20}55\% & \cellcolor{gray!20}43.4 & \cellcolor{gray!20}53.7 & \cellcolor{gray!20}0.43 & \cellcolor{gray!20}79\% & \cellcolor{gray!20}45.1 & \cellcolor{gray!20}61.7 \\
&                      & \cellcolor{gray!20}Ours   & \cellcolor{gray!20}1 & \cellcolor{gray!20}0.56 & \cellcolor{gray!20}0.47 &\cellcolor{gray!20} 0.28 & \cellcolor{gray!20}55\% (+0\%) & \cellcolor{gray!20}49.6 & \cellcolor{gray!20}58.3 & \cellcolor{gray!20}0.29 & \cellcolor{gray!20}79\% & \cellcolor{gray!20}36.7 & \cellcolor{gray!20}59.3 \\
\cmidrule(lr){2-14}
& \multirow{3}{*}{20\%} & GA     & 6 & 0.73 & 0.52 & 0.50 & 49\% & 31.5 & 48.6 & 0.73 & 79\% & 27.0 & 53.1 \\
&                      & SimNPO & 6 & 0.72 & 0.51 & 0.57 & 40\% & 28.4 & 47.0 & 0.77 & 77\% & 24.3 & 51.5 \\
&                      & \cellcolor{gray!20}Ours   & \cellcolor{gray!20}1 & \cellcolor{gray!20}0.56 & \cellcolor{gray!20}0.47 & \cellcolor{gray!20}0.28 & \cellcolor{gray!20}57\% (+2\%) & \cellcolor{gray!20}48.5 & \cellcolor{gray!20}58.3 & \cellcolor{gray!20}0.30 & \cellcolor{gray!20}80\% & \cellcolor{gray!20}37.6 & \cellcolor{gray!20}58.2 \\
\cmidrule(lr){2-14}
& \multirow{3}{*}{30\%} & GA     & 5 & 0.71 & 0.51 & 1.00 & 28\% & 19.2 & 42.3 & 1.10 & 16\% & 10.9 & 35.4 \\
&                      & SimNPO & 5 & 0.71 & 0.50 & 0.73 & 28\% & 17.9 & 40.9 & 1.11 & 17\% & 10.8 & 34.8 \\
&                      & \cellcolor{gray!20}Ours   & \cellcolor{gray!20}1 & \cellcolor{gray!20}0.55 & \cellcolor{gray!20}0.45 & \cellcolor{gray!20}0.29 & \cellcolor{gray!20}53\% (-2\%) & \cellcolor{gray!20}47.9 & \cellcolor{gray!20}56.6 & \cellcolor{gray!20}0.29 & \cellcolor{gray!20}86\% & \cellcolor{gray!20}41.1 & \cellcolor{gray!20}62.0 \\

\midrule
%================ Qwen ==================
\multirow{10}{*}{Qwen-7B}
& --- & --- & --- & 0.87 & 0.96 & 0.25 & 57\% & 50.0 & 57.7 & 0.28 & 82\% & 48.2 & 68.1 \\
\cmidrule(lr){2-14}
& \multirow{3}{*}{10\%} & \cellcolor{gray!20}GA     & \cellcolor{gray!20}6 & \cellcolor{gray!20}0.51 & \cellcolor{gray!20}0.47 & \cellcolor{gray!20}0.37 & \cellcolor{gray!20}54\% & \cellcolor{gray!20}37.5 & \cellcolor{gray!20}48.0 & \cellcolor{gray!20}0.39 & \cellcolor{gray!20}80\% & \cellcolor{gray!20}46.3 & \cellcolor{gray!20}58.1\\
&                      & \cellcolor{gray!20}SimNPO & \cellcolor{gray!20}5 & \cellcolor{gray!20}0.54 & \cellcolor{gray!20}0.53 & \cellcolor{gray!20}0.33 & \cellcolor{gray!20}55\% & \cellcolor{gray!20}41.0 & \cellcolor{gray!20}50.0 & \cellcolor{gray!20}0.39 & \cellcolor{gray!20}81\% & \cellcolor{gray!20}47.2 & \cellcolor{gray!20}59.3\\
&                      & \cellcolor{gray!20}Ours   & \cellcolor{gray!20}1 & \cellcolor{gray!20}0.60 & \cellcolor{gray!20}0.47 & \cellcolor{gray!20}0.25 & \cellcolor{gray!20}56\% (-1\%) & \cellcolor{gray!20}49.2 & \cellcolor{gray!20}57.3 & \cellcolor{gray!20}0.28 & \cellcolor{gray!20}85\% & \cellcolor{gray!20}48.0 & \cellcolor{gray!20}64.8\\
\cmidrule(lr){2-14}
& \multirow{3}{*}{20\%} & GA     & 4 & 0.61 & 0.51 & 0.43 & 30\% & 37.8 & 47.4 & 0.77 & 55\% & 29.4 & 50.9 \\
&                      & SimNPO & 4 & 0.60 & 0.50 & 0.45 & 20\% & 36.4 & 45.9 & 0.79 & 42\% & 28.8 & 50.1 \\
&                      & \cellcolor{gray!20}Ours   & \cellcolor{gray!20}1 & \cellcolor{gray!20}0.57 & \cellcolor{gray!20}0.51 & \cellcolor{gray!20}0.25 & \cellcolor{gray!20}53\% (-4\%) & \cellcolor{gray!20}46.9 & \cellcolor{gray!20}54.7 & \cellcolor{gray!20}0.32 & \cellcolor{gray!20}83\% & \cellcolor{gray!20}46.6 & \cellcolor{gray!20}61.7\\
\cmidrule(lr){2-14}
& \multirow{3}{*}{30\%} & GA     & 3 & 0.64 & 0.50 & 0.49 & 13\% & 31.8 & 44.2 & 0.80 & 40\% & 24.0 & 41.0\\
&                      & SimNPO & 3 & 0.64 & 0.50 & 0.50 & 10\% & 34.9 & 44.8 & 0.88 & 45\% & 23.4 & 40.1\\
&                      & \cellcolor{gray!20}Ours   & \cellcolor{gray!20}1 & \cellcolor{gray!20}0.56 & \cellcolor{gray!20}0.49 & \cellcolor{gray!20}0.27 & \cellcolor{gray!20}55\% (-2\%) & \cellcolor{gray!20}46.2 & \cellcolor{gray!20} 55.0 & \cellcolor{gray!20}0.34 & \cellcolor{gray!20}86\% & \cellcolor{gray!20}49.2 & \cellcolor{gray!20}63.0\\

\bottomrule
\end{tabular}}
\label{tab:result}

\vskip -12pt
\end{table*}
}

\subsubsection{Unlearning Baselines and Configurations}
We compare our method against two of the most recent and representative LLM unlearning approaches, which SALAD~\cite{wang2025salad} used. 
\textbf{Gradient Ascent (GA)}~\cite{yao2024large} maximizes the cross-entropy loss on forget samples:
\begin{equation}
\mathcal{L}_{\text{GA}}(\theta) =
- \mathbb{E}_{(x,y)\in \mathcal{D}_f} \left[ -\log \pi_\theta(y \mid x) \right].
\end{equation}
\textbf{SimNPO}~\cite{fan2024simplicity} introduces a smooth penalty function to improve stability and avoid collapse of GA:
\begin{equation}
\mathcal{L}_{\text{SimNPO}}(\theta) =
\mathbb{E}_{(x,y)\in \mathcal{D}_f} \left[
-\tfrac{2}{\beta} \log \sigma \!\left(
-\tfrac{\beta}{|y|} \log \pi_\theta(y \mid x) - \gamma
\right)
\right].
\end{equation}
To ensure fairness, we follow the reward-margin settings in~\cite{fan2024simplicity} to set $\beta=2.5$ and $\gamma=0$.
Our \textbf{FiFSL} further introduces an adaptive floor parameter ($L_\text{min}$), chosen as 0.35, which corresponds to the validation loss after the first fine-tuning epoch.
In addition, we explore a broader hyperparameter space by using Llama-7B as an example, varying $\beta$ between 1.5 and 4 and $\gamma$ from 0 to 2.5.
Consistent with the findings of~\cite{fan2024simplicity}, the configuration $\beta=2.5, \gamma=0$ achieves the best performance.
We build our syntax-preserving dataset guided by~\cite{10458102, 1620780, palnitkar2003verilog}.
Unlearning is performed using a learning rate of $2\times10^{-6}$ until MinK++ converges near $0.5$, ensuring effective forgetting without excessive utility degradation.

\subsection{Experimental Results}
\subsubsection{Forgetting Effectiveness vs. Utility Preservation}
We compare our approach against GA and SimNPO in terms of forget quality and model utility.
Fig.~\ref{fig:Results} shows results across different forget sets and benchmark models.
We adopt MinK++ as the indicator of forget quality and Pass@1 as the measure of model utility, since they directly capture the trade-off between eliminating memorized knowledge and retaining downstream RTL code generation reliability.
Both values are normalized for comparability.
An ideal unlearning method aspires to the point (1.0, 1.0) in the forget–utility space, denoting perfect forgetting without utility loss.
Our method consistently lies closest to this ultimate goal across all models and forget-set sizes.
In particular, it enables forgetting up to 30\% of the training corpus across coder LLMs, while conventional methods fail beyond 10\% (i.e., a 3$\times$ larger forget set).
For Llama-3-8B, our method achieves up to 20\% forgetting, which we attribute to model-specific architectural constraints, as this model is not explicitly optimized for code understanding and generation. 
Nevertheless, this still outperforms conventional methods, demonstrating the robustness and broad applicability of our approach.
Importantly, our method preserves the highest model utility among all methods: at 30\% forgetting, it retains up to 14.3$\times$ higher Pass@1 than baselines, whereas GA and SimNPO exhibit sharp utility degradation. 
Concrete values across all metrics are reported in Table~\ref{tab:result}.
Moreover, our method accomplishes near-complete forgetting within \textit{a single epoch}, drastically reducing unlearning cost, while GA and SimNPO require 4$\sim$7 epochs and underperform, particularly on larger forget sets.

{\renewcommand{\arraystretch}{0.9}
\begin{table}[!t]
\centering
% \vskip -5pt
\caption{Performance on VerilogEval and Rtllm Benchmarks.}
\vskip -6pt
\scriptsize
\small
\resizebox{\linewidth}{!}{
\begin{tabular}{l c | c c | c c}
\toprule
\multirow{2}{*}{Model} & 
\multirow{2}{*}{Forget sets} & 
\multicolumn{2}{c|}{\textbf{VerilogEval}~\cite{liu2023verilogeval}} & 
\multicolumn{2}{c}{\textbf{Rtllm}~\cite{lu2024rtllm}} \\
\cmidrule(lr){3-4}\cmidrule(lr){5-6}
& & Pass@1 $\uparrow$ & Pass@5 $\uparrow$ 
& Pass@1 $\uparrow$ & Pass@5 $\uparrow$ \\
\midrule

% ------ fill rows here ------

\multirow{4}{*}{Llama-7B} 
& ---     & 30\%  & 39\%  & 34\%  & 40\%  \\
& 10\%   & 30\%  & 38\%  & 32\%  &  40\% \\
& 20\%   & 29\%  & 39\%  & 32\%  &  40\% \\
& 30\%   & 28\%  & 37\%  & 30\%  &  38\% \\
\midrule

% ------------------- Llama-8B -------------------
\multirow{4}{*}{Llama-8B} 
& ---     & 35\%  & 46\%  & 36\%  & 44\%  \\
& 10\%   & 31\%  & 45\%  & 34\%  &  44\% \\
& 20\%   & 31\%  & 43\%  & 38\%  &  42\% \\
& 30\%   & 28\%  & 38\%  & 30\%  &  36\% \\
\midrule

% ------------------- DeepSeek-7B -------------------
\multirow{4}{*}{DeepSeek-7B} 
& ---     & 35\%  & 44\%  & 40\%  & 44\%  \\
& 10\%   & 37\%  &  45\% & 40\%  &  44\% \\
& 20\%   & 38\%  &  44\% & 42\%  &  44\% \\
& 30\%   & 33\%  & 42\%  & 38\%  &  42\% \\
\midrule

% ------------------- Qwen-7B -------------------
\multirow{4}{*}{Qwen-7B} 
& ---     & 36\%  & 49\%  & 40\%  &  48\% \\
& 10\%   &  36\% & 48\%  & 40\%  &  48\% \\
& 20\%   & 34\%  & 48\%  & 40\%  &  46\% \\
& 30\%   & 33\%  & 44\%  & 36\%  & 44\%  \\
\bottomrule
\end{tabular}}
\label{tab:passk}
\vskip -6pt
\end{table}
}

\subsubsection{Generalization Ability and Scalability}
Table~\ref{tab:result} reports all evaluation metrics, where ``epoch'' denotes the number of training epochs required to reach ideal forget quality, with $\downarrow$ ($\uparrow$) indicating that lower (higher) values are better.
Our method surpasses in all metrics.
We further assess generalization on unseen, simple code generation tasks, shown as generalization ability columns. 
The shaded rows highlight that the unlearned models maintain Pass@1 close to their fine-tuned baselines, confirming that forgetting does not compromise general-purpose generation ability.
In addition, the proposed approach sustains high BLEU and chrF scores across these tasks, demonstrating that effective unlearning can be achieved without sacrificing broad code generation performance.
To further validate domain-level utility preservation, we additionally evaluate our method on VerilogEval~\cite{liu2023verilogeval} and Rtllm~\cite{lu2024rtllm} before and after unlearning. As shown in Table~\ref{tab:passk}, the unlearned models retain competitive performance across both benchmarks, confirming that the proposed method does not compromise downstream RTL generation utility.

\subsubsection{Summarization}
Overall, our domain-specific method supports up to 3$\times$ larger forget sets than existing methods (i.e., GA and SimNPO used by SALAD~\cite{wang2025salad} without domain adaptation), achieves effective forgetting in only one epoch, and sustains both validation and generalization performance (Table~\ref{tab:unlearning_comparison}). 
These results show that it offers near-ideal unlearning: strong forgetting quality, preserved model utility, broad generalization, and exceptional efficiency, making it a scalable and practical solution for reliable code-specialized LLMs.

\begin{table}[t]
\vskip -5pt
\centering
\caption{Overall comparison with baselines.}
\vskip -5pt
\begin{tabular}{|l|c|c|c|}
\hline
\textbf{Method} & \textbf{Forget Sets} & \textbf{Epochs} & \textbf{Utility} \\ \hline
GA     & $1\times$      & 4$\sim$7 & $\times$ \\ 
SimNPO & $1\times$      & 4$\sim$7 & $\times$ \\

\rowcolor{gray!20}
Ours   & $2\times$$\sim$$3\times$ & 1 & \checkmark \\ \hline
\end{tabular}
\vskip -13pt
\label{tab:unlearning_comparison}
\end{table}

\section{Conclusion}
\label{sec:conclusion}

We present the first domain-specific unlearning framework for RTL code generation, addressing critical reliability challenges in hardware design automation. 
By combining a syntax-preserving strategy with the proposed FiFSL objective, our approach enables precise and efficient forgetting while maintaining syntactic validity and functional reliability. 
Experiments show that it supports up to 3$\times$ larger forget sets with only one training epoch, significantly outperforming prior methods. 
This work demonstrates a practical and scalable path toward reliable hardware code generation with LLMs.

\bibliographystyle{IEEEtran}
\bibliography{biography}

\end{document}